\documentclass[letterpaper]{article} 
\usepackage{aaai24}  
\usepackage{times}  
\usepackage{helvet}  
\usepackage{courier}  
\usepackage[hyphens]{url}  
\usepackage{graphicx} 
\usepackage{natbib}  
\usepackage{caption} 
\usepackage{algorithm}
\usepackage{algorithmic}
\usepackage{graphicx}
\usepackage{amsmath}
\usepackage{amssymb}
\usepackage{booktabs}
\usepackage{multirow}
\usepackage{makecell}

\usepackage{newfloat}
\usepackage{listings}
\DeclareCaptionStyle{ruled}{labelfont=normalfont,labelsep=colon,strut=off} 
\floatstyle{ruled}
\newfloat{listing}{tb}{lst}{}
\floatname{listing}{Listing}
\title{DenseMP: Unsupervised Dense Pre-training for Few-shot Medical Image Segmentation}
\author{
    Jason Zhaoxin Fan\textsuperscript{\rm 1},
    Puquan Pan\textsuperscript{\rm 2},
    Zeren Zhang\textsuperscript{\rm 2},
    Ce Chen\textsuperscript{\rm 2}\\
    Tianyang Wang\textsuperscript{\rm 2},
    Siyang Zheng\textsuperscript{\rm 1},
    Min Xu\textsuperscript{\rm 1}\thanks{Corresponding author: mxu1@cs.cmu.edu}\\
    \textsuperscript{\rm 1}Carnegie Mellon University, 
    \textsuperscript{\rm 2}Austin Peay State University,\\
}
\affiliations{\{zhaoxinf, siyangz,mxu1\}@andrew.cmu.edu,\{msppqkasri, antonio.chan.cc, zzeren030\}@gmail.com, toseattle@siu.edu\\
}

\usepackage{bibentry}

\begin{document}

\maketitle

\begin{abstract}
Few-shot medical image semantic segmentation is of paramount importance in the domain of medical image analysis. However, existing methodologies grapple with the challenge of data scarcity during the training phase, leading to over-fitting. To mitigate this issue, we introduce a novel Unsupervised Dense Few-shot Medical Image Segmentation Model Training Pipeline (DenseMP) that capitalizes on unsupervised dense pre-training. DenseMP is composed of two distinct stages: (1) segmentation-aware dense contrastive pre-training, and (2) few-shot-aware superpixel guided dense pre-training. These stages collaboratively yield a pre-trained initial model specifically designed for few-shot medical image segmentation, which can subsequently be fine-tuned on the target dataset. Our proposed pipeline significantly enhances the performance of the widely recognized few-shot segmentation model, PA-Net, achieving state-of-the-art results on the Abd-CT and Abd-MRI datasets. Code will be released after acceptance.
\end{abstract}

\section{Introduction}

Medical image segmentation plays a pivotal role in the field of medical image analysis, as it facilitates the precise identification and delineation of anatomical structures and pathological regions within the images \cite{zhang2018geometric,zhang2013challenges}. This, in turn, enables accurate disease diagnosis \cite{Chen2019Chronic}, treatment planning \cite{trofimov2007radiotherapy}, and monitoring of therapeutic interventions \cite{wright2020remote}. Among various approaches, deep learning-based methods have emerged as the most prevalent solutions, owing to their exceptional performance in extracting complex patterns and features from the data \cite{pham2000survey,sharma2010automated,hesamian2019deep}.

Traditionally, such deep learning models rely heavily on large-scale, well-annotated datasets for supervised training. However, this requirement poses a significant challenge, as acquiring and annotating medical image data is a labor-intensive and time-consuming process, often involving domain experts. Although weakly supervised \cite{girum2020fast,roth2021going,xu2014weakly} and unsupervised methods \cite{aganj2018unsupervised,chen2020unsupervised,perone2019unsupervised} have been proposed to alleviate this issue, they come with their own set of shortcomings. For instance, weakly supervised methods might suffer from noisy labels and suboptimal performance, while unsupervised methods may struggle with capturing high-level semantic information pertinent to the specific segmentation task. In light of these limitations, few-shot segmentation appears as an attractive alternative, as it aims to learn effective models with a minimal number of annotated examples. This approach holds great promise for addressing the challenges associated with data scarcity in medical image segmentation and offers a viable direction for future research.

\begin{figure}[t]
	\centering
	\includegraphics[width=0.9\linewidth]{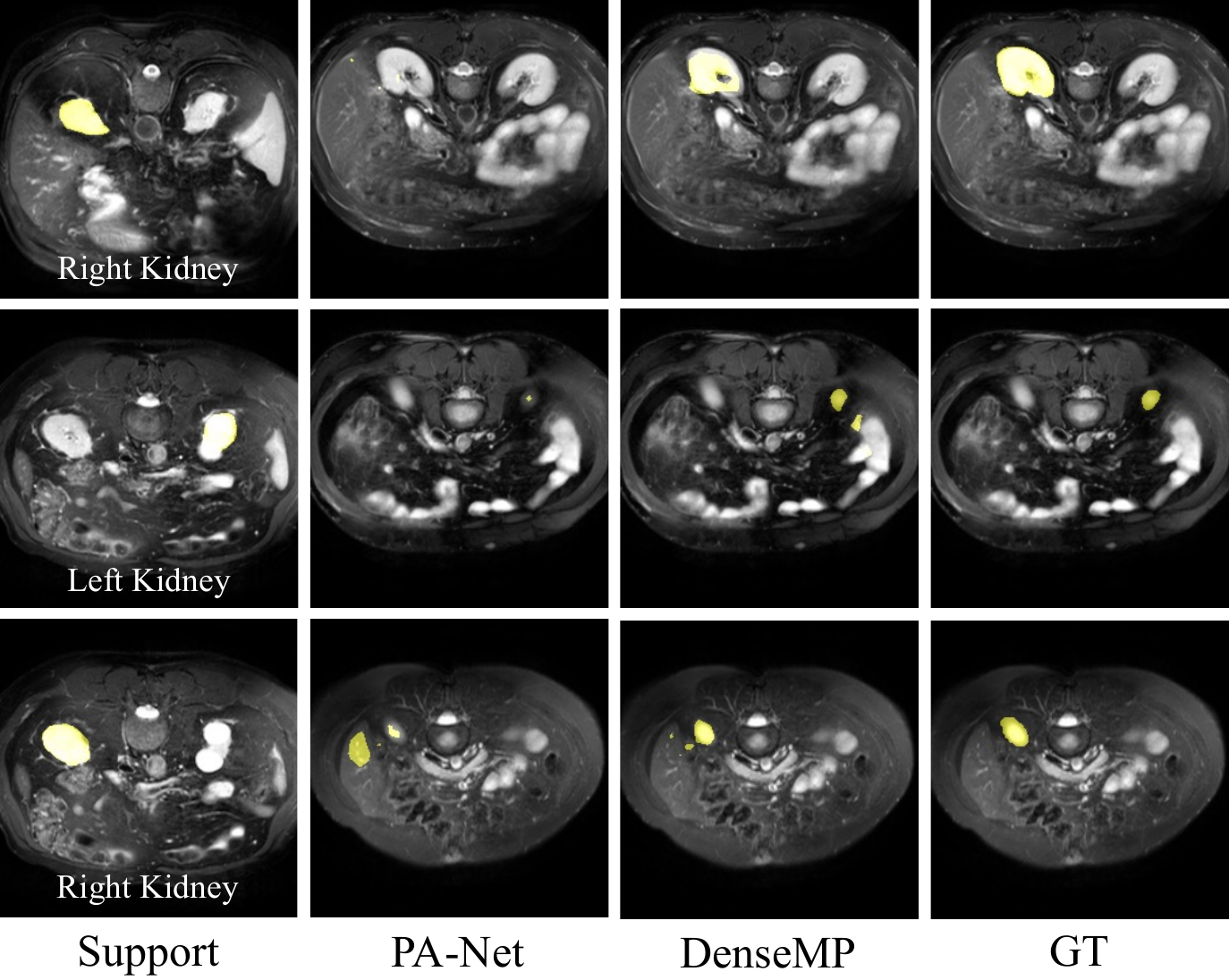}
	\caption{An example of few-shot medical image segmentation results before and after unsupervised dense pre-training. PA-Net is our baseline without adopting unsupervised dense pre-training, and DenseMP is our method adopting unsupervised dense pre-training.}
	\label{fig:intro}
\end{figure}

In the few-shot segmentation paradigm, a model is trained on an extensive dataset comprising images and their corresponding ground truth masks. During the testing phase, the model is tasked with segmenting novel classes in a query image by referencing support images and their associated ground truth masks. Substantial advancements have been made in the few-shot medical image segmentation domain \cite{rakelly2018conditional,shaban2017one,dong2018few,siam2019adaptive,wang2019panet,zhang2020sg,zhang2019pyramid,yan2019dual,hu2019attention,fan2020graph}, and several efforts have been dedicated to adapting these methods specifically for medical image segmentation \cite{roy2020squeeze,ouyang2020self}. Nevertheless, the prevailing training approach for few-shot medical image segmentation still demands a considerable volume of annotated data. For instance, in the case of segmenting brain tumors, acquiring a large dataset with expert-annotated ground truth masks for various tumor subregions is challenging, as it requires the collaboration of experienced radiologists or neurosurgeons. The time-consuming and labor-intensive nature of this process exacerbates the issue. Consequently, a trained model tends to be susceptible to over-fitting observed classes. Consequently, a trained model tends to be susceptible to over-fitting observed classes. In practical applications, this over-fitting issue can severely undermine the generalization capabilities of few-shot medical image segmentation methods, leading to inaccurate segmentations and potentially detrimental clinical decisions. Thus, despite the potential advantages of few-shot learning, these over-fitting challenges still limit the practical utility of current approaches in real-world medical settings.

Recognizing the limitations of existing few-shot medical image segmentation methods, particularly the over-fitting issue and its implications on practical applications, we propose an innovative solution to address these challenges: DenseMP, an unsupervised dense few-shot medical image segmentation model training pipeline. The DenseMP pipeline is specifically designed to tackle the over-fitting issue through two complementary stages. The first stage, the segmentation-aware dense contrastive pre-training, concentrates on learning representative features optimally suited for the medical image segmentation task. By employing a dense contrastive training scheme, the model effectively captures intricate and diverse patterns in medical images, enhancing its segmentation capabilities. The second stage, the few-shot-aware superpixel guided dense pre-training, focuses on addressing the unique challenges associated with few-shot learning. This stage utilizes superpixels as pseudo labels and simulates the process of few-shot segmentation, fostering the acquisition of task-oriented features and enabling the model to adapt more effectively to limited labeled data. Through the combination of these two distinct yet synergistic stages, our DenseMP pipeline is tailored specifically for few-shot medical image segmentation, effectively mitigating the over-fitting issue and enhancing the practical utility of such models in real-world medical scenarios. 

To validate the efficacy of our proposed DenseMP pipeline, we apply it to PA-Net \cite{wang2019panet}, a well-established few-shot segmentation model for natural images. Experimental results on the Abd-CT \cite{landman2015miccai} and Abd-MRI \cite{kavur2021chaos} datasets demonstrate that our DenseMP significantly enhances the performance of PA-Net. Importantly, our approach outperforms existing pre-training methods, such as SimCLR \cite{chen2020simple}, as it is specifically tailored for few-shot medical image segmentation. As a result, our method achieves a new state-of-the-art in this domain.

Our contributions can be summarized as follows:

\begin{itemize}
    \item We propose DenseMP, a novel unsupervised pre-training pipeline tailored for few-shot medical image segmentation, which substantially improves the performance of the PA-Net baseline and outperforms existing pre-training methods like SimCLR.
    \item We introduce two key components of the unsupervised pre-training: the segmentation-aware dense contrastive pre-training stage and the few-shot-aware superpixel guided dense pre-training stage.
    \item We conduct extensive evaluations of our method on the well-known Abd-CT and Abd-MRI datasets, demonstrating its superiority over existing state-of-the-art pre-training approaches.
\end{itemize}

\section{Related work}

\subsection{Traditional semantic segmentation}

Semantic segmentation aims to assign a classification label to each pixel in an image. Numerous approaches have been proposed for this task, starting with the pioneering work of FCN \cite{long2015fully}, which employed a fully convolutional network for semantic segmentation. To enhance its performance, later methods, such as CRF/MRF \cite{chen2014semantic, liu2015semantic, zheng2015conditional}, proposed various post-processing modules to refine the results. Subsequently, numerous encoder-decoder structures \cite{badrinarayanan2015deep, noh2015learning, ronneberger2015u} were proposed, which now dominate state-of-the-art semantic segmentation methods. Pre-training techniques have also gained considerable attention for improving network structures \cite{long2015fully}, with dense contrastive learning \cite{wang2021dense} standing out as a leading approach due to its annotation-free pre-training process. In the field of medical image segmentation, specialized methods have been proposed to address the unique challenges associated with this domain. U-Net \cite{ronneberger2015u} is one of the most widely used network architectures for medical image segmentation. Following this, numerous U-Net variants, such as Res-UNet \cite{xiao2018weighted}, Dense-UNet \cite{li2018h}, U-Net++ \cite{zhou2018unet++}, and UNet3+ \cite{huang2020unet}, were introduced to learn more robust features. For a comprehensive overview of semantic segmentation, we kindly refer readers to the survey by \cite{ulku2022survey}.

While significant progress has been made in both natural image segmentation and medical image segmentation, these methods predominantly rely on large samples, i.e., they require a substantial amount of annotated data for each new class, necessitating time-consuming training. In contrast, this paper focuses on the more advanced and challenging task of few-shot medical image segmentation, which aims to achieve high segmentation performance when only a limited number of annotated examples are available.

\subsection{Few-shot semantic segmentation}

Typically, few-shot learning focuses on data, model, and algorithm \cite{wang2020generalizing}.  Prototypical networks \cite{snell2017prototypical} is a classic work that learns a prototype from the support image for each class in image classification. Then, Dong et al.\cite{dong2018few} extend the prototypical networks into few-shot image segmentation by using a global average pooling layer to learn prototypes from masked regions. Then, PANet \cite{wang2019panet} proposes a prototype alignment network to further improve the performance. After that, many following works \cite{liu2020crnet,tian2020prior,li2021adaptive} are designed in the literature from the perspective of learning to extract better prototypes, learning priors, building better network architectures, and so on. Recently, \cite{yang2021mining} add an additional mining branch which exploits novel classes via transferable sub-clusters and a new rectification technique to enforce more stable prototypes.
\cite{wu2021learning} introduce the concept of meta-class, which is the meta information shareable among all classes.
Though promising, the performance of directly migrating these methods to few-shot medical segmentation is not satisfactory. Therefore, works tailored for medical images are proposed \cite{roy2020squeeze, ouyang2020self}. However, there is still much room for their performance improvement.

We hypothesize that the difficulty in adapting existing few-shot natural image segmentation models to medical images arises from the relatively small scale of available medical image segmentation datasets. This limitation can lead to overfitting problems in few-shot medical image segmentation, as the network struggles to learn general features for unseen classes. In this paper, we propose the novel DenseMP method to address this issue from the perspective of dense pre-training, aiming to improve few-shot medical image segmentation performance by leveraging more expressive feature representations. 

\subsection{Unsupervised pre-training}
Modern pre-training techniques have been shown to significantly improve performance on downstream tasks across various domains, including computer vision \cite{He_2019_ICCV}, natural language processing \cite{devlin2018bert}, and cross-domain applications \cite{chen2020uniter, yang2021tap}. Recently, unsupervised pre-training has gained popularity due to its ability to leverage models without requiring ground-truth labels. Most existing unsupervised pre-training models adopt a contrastive learning pipeline, wherein a loss is defined to bring positive samples closer and push negative samples apart. Consequently, the definition of positive and negative samples is crucial. In the natural language processing domain, temporally close sentences or tokens \cite{logeswaran2018efficient, wang2021cline} are often considered as positive samples, while randomly selected sentences are treated as negative samples. In the image processing domain, methods like MoCo \cite{he2020momentum}, SimCLR \cite{chen2020simple}, and masked autoencoders (MAEs) \cite{he2022masked} have been proposed. MoCo and SimCLR employ random augmentations to generate positive samples of an image, while any other image serves as a negative sample. In contrast, MAEs focus on masking random patches of the input image and reconstructing the missing pixels, leading to a scalable self-supervised learning approach for computer vision tasks. Although MoCo and SimCLR demonstrate promising results, they are global pre-training methods that are more suitable for downstream tasks involving global information, such as classification, rather than dense tasks like object detection and segmentation. To benefit these dense tasks, methods like \cite{wang2021dense} propose dense contrastive learning for pre-training deep models. However, dense contrastive learning has not yet been explored in the context of medical image segmentation.

In this paper, we present the first application of dense contrastive learning to medical image segmentation. Specifically, we propose a segmentation-aware dense contrastive pre-training module based on dense contrastive learning to pre-train the backbone of a few-shot medical image segmentation model. This approach aims to leverage the advantages of dense contrastive learning for enhancing the performance of few-shot medical image segmentation tasks, in accordance with the high standards set by top-tier conferences and journals such as ICML and Nature.

\section{Method}

In this section, we first provide a formal definition of the few-shot medical image segmentation problem. Subsequently, we present an overview of our proposed method, delineating its primary components. Finally, we elaborate on the proposed Segmentation-aware Dense Contrastive Pre-training and the Few-shot-aware Superpixel Guided Dense Pre-training in detail.

\begin{figure*}[ht]
	\centering
	\includegraphics[width=0.99\textwidth]{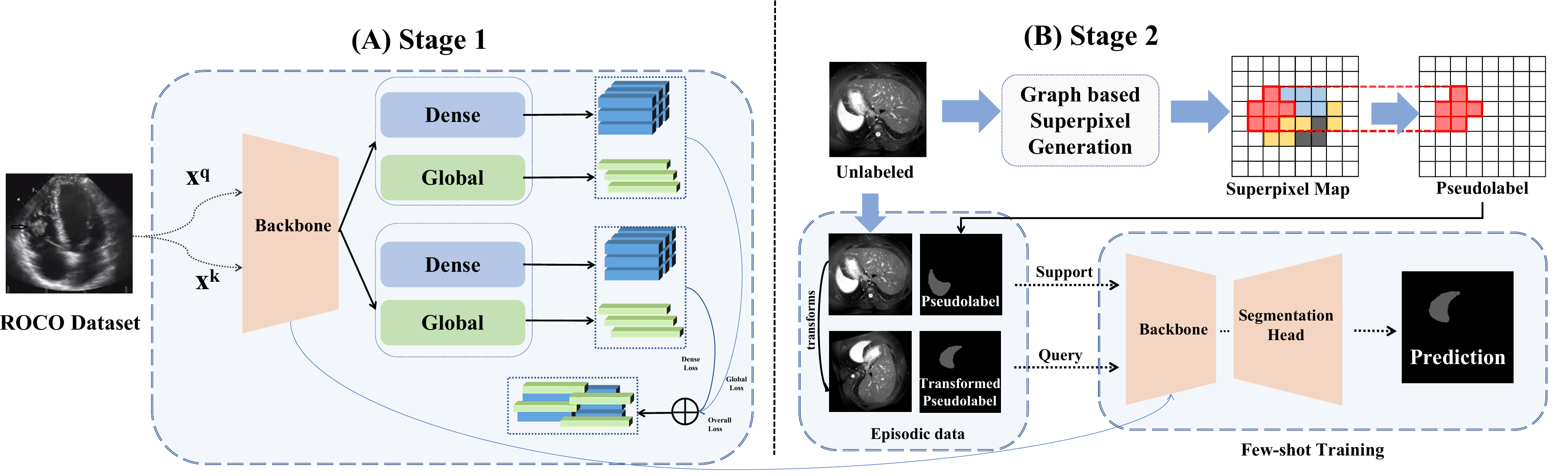}
	\caption{Overview of DenseMP, which consists of two stages: (A) segmentation-aware dense contrastive pre-training and (B) few-shot-aware superpixel guided dense pre-training.}
	\label{fig:overview}
\end{figure*}

\subsection{Problem Formulation}
We formulate the few-shot medical image segmentation task as an episodic training and testing problem, following the paradigm established by \cite{ouyang2020self}. An episode encompasses $K$ support images $I_s$, their corresponding ground-truth masks $M_s$ for $N$ classes, and a query image $I_q$. The primary objective is to exploit the information present in $I_s$, $I_q$, and $M_s$ to predict the segmentation mask $M_q$, a scenario referred to as an N-way-K-shot problem. Generally, the condition $C_{tr} \cap C_{te}=\emptyset$ holds, indicating that during testing, the model is not required to be pre-trained or fine-tuned on the new class using a large amount of labeled data. Instead, a limited number of labeled support images are leveraged to facilitate the segmentation process. In the context of medical image segmentation, the majority of works concentrate on the 1-way 1-shot case \cite{roy2020squeeze,ouyang2020self}. We adhere to this setting as well, formulating the task as follows:

\begin{equation}
\hat M_{q}=f(\{I_s,I_q\},M_s),
\end{equation}

where $f$ represents a trainable model, such as a deep neural network.

\subsection{Overview}
 We adopt the PANet model \cite{wang2019panet} as our task model, which comprises a backbone component for feature extraction from support and query images, and a segmentation head for generating segmentation masks. Let $I^s$ and $I^q$ denote the support and query images, respectively. The backbone component extracts feature maps $\mathcal{F}^s$ and $\mathcal{F}^q$:

$$\mathcal{F}^s = \text{Backbone}(I^s)$$
$$\mathcal{F}^q = \text{Backbone}(I^q)$$

The adaptive local prototype pooling technique, introduced by \cite{ouyang2020self}, is employed in the segmentation head to extract prototypes $\mathcal{P}$ of the target classes:

$$\mathcal{P} = \text{AdaptivePooling}(\mathcal{F}^s)$$

A similarity-based segmentation strategy is then applied to compare the prototypes with the query feature maps, resulting in a segmentation mask $M$:

$$M = \text{Similarity}(\mathcal{P}, \mathcal{F}^q)$$.

Given the limited size of medical image datasets, the network struggles to learn adequate general features for few-shot segmentation, particularly for unobserved classes. To overcome this limitation, we propose DenseMP, an unsupervised dense pre-training method for medical images. As depicted in Fig. \ref{fig:overview}, our proposed method, DenseMP, encompasses two distinct stages: (1) Segmentation-aware Dense Contrastive Pre-training, and (2) Few-shot-aware Superpixel Guided Dense Pre-training. In the following sections, we offer a comprehensive explanation of each stage.

\subsection{Stage 1: Segmentation-aware Dense Contrastive Pre-training}
In this section, we present our approach for unsupervised pre-training of the backbone network. Recently, numerous works have been proposed for unsupervised pre-training of natural images, such as MoCo \cite{he2020momentum}. The common pipeline for these unsupervised methods involves generating an augmented view of each image as a positive sample, while treating other distinct images as negative samples. Subsequently, contrastive learning is employed to train the network, removing the need for ground truth to guide the training process. Although these pre-trained models have demonstrated satisfactory performance in downstream tasks, we argue that they are not ideally suited for segmentation tasks. This is because they rely solely on the global feature vector for contrastive learning, potentially causing the learned feature map to lose local details, resulting in a sparse rather than dense information representation. However, dense local information is crucial for dense segmentation tasks. Consequently, it is imperative that the backbone learns such dense local information during pre-training. A direct approach would be to pre-train the backbone using fully supervised techniques, but this is impractical due to the difficulty of procuring a large-scale densely annotated dataset. As a result, we propose employing dense pre-training \cite{wang2021dense} technology to pre-train our backbone in this work, which we refer to as segmentation-aware dense contrastive pre-training.

Given a medical image $I$, we first apply various random augmentations to generate $K$ distinct views, denoted as \{$I_{v1}$, $I_{v2}, \ldots, I_{vk}$\}. These views are fed into the backbone network to produce feature maps \{$F_{v1}, F_{v2}, \ldots, F_{vk}$\}. Assuming their dimensions are $S_W \times S_H \times C$, a projection head is utilized to project them into feature maps with dimensions $S \times S \times C$. For each feature map, $S \times S$ feature vectors can be obtained, denoted as encoded keys \{$t_0, t_1, \ldots, t_{ss}$\}. For each key $t$, its negative key $t_-$ can be easily defined as keys randomly selected from other images (excluding other views). To identify the positive keys, an adaptive pool is employed to project \{$F_{v1}, F_{v2}, \ldots, F_{vk}$\} into feature maps \{$F_{a1}, F_{a2}, \ldots, F_{ak}$\} with dimensions $S \times S$, allowing us to obtain alignment vectors \{$a_0, a_1, \ldots, a_{ss}$\}. Then, the similarity between $a$ and all other alignment vectors in the other views of the same image is computed, with the most similar one's corresponding encoded key defined as its positive key $t_+$. Consequently, the dense contrastive loss can be defined as:

\begin{equation}
    \mathcal{L}_t = \frac {1}{S^2} \sum_{s} - \log\left(\frac {e^{(t \cdot t_+) / \tau}}{e^{(t \cdot t_+)} + \sum_{t_-} e^{(t \cdot t_-) / \tau}}\right),
\end{equation}
where $\tau$ denotes a temperature hyperparameter.

To preserve the network's capacity to represent global information, we also obtain the global feature vector of each view, defined as \{$g_0$, $g_1$, ..., $g_k$\}. For each $g$, the positive key $g_+$ is the global vector of another view of the same image, while the negative key is the global vector of a different image. Thus, the global contrastive loss can be defined as:

\begin{equation}
    \mathcal{L}_g = - \log\left(\frac {e^{(g \cdot g_+) / \tau}}{e^{(g \cdot g_+)} + \sum_{g_-} e^{(g \cdot g_-) / \tau}}\right).
\end{equation}

Overall, the total loss for our method can be formulated as:

\begin{equation}
    \mathcal{L} = (1 - \lambda) \mathcal{L}_g + \lambda \mathcal{L}_t,
\end{equation}

where $\lambda$ is a balance term.

The proposed Segmentation-aware Dense Contrastive Pre-training offers several advantages for few-shot medical image segmentation in the context of our approach. First, it enables the backbone network to learn rich, dense local information during pre-training, which is essential for capturing fine-grained details in segmentation tasks. This is particularly crucial in medical image segmentation, where accurate delineation of boundaries and structures can significantly impact diagnosis and treatment planning. Second, the unsupervised nature of the pre-training process circumvents the need for large-scale, densely annotated datasets, which are often difficult to procure in the medical domain due to privacy restrictions and the labor-intensive nature of manual annotation. Lastly, by incorporating both dense local and global feature learning, the segmentation-aware dense contrastive pre-training is capable of effectively leveraging multi-scale information, leading to enhanced generalization and performance in few-shot medical image segmentation tasks. This approach aligns well with our objective of building a robust and efficient segmentation model with limited annotated data.

\begin{table*}[ht]
    \centering
    \caption{Comparison with sota few-shot medical image segmentation methods on setting 1.}
    \label{table:setting1}
    \begin{tabular}{l|cccc|c|cccc|c}
        \toprule
        & \multicolumn{5}{c}{Abd-CT Dice} & \multicolumn{5}{c}{Abd-MRI Dice} \\
        \cmidrule(lr){2-6} \cmidrule(lr){7-11}
        Method & RK & LK & Liver & Spleen & Mean & RK & LK & Liver & Spleen & Mean \\
        \midrule
        Vanilla PANet & 21.19 & 20.67 & 49.55 & 36.04 & 31.86 & 32.19 & 30.99 & 50.40 & 40.58 & 38.53 \\
        SE-Net & 12.51 & 24.42 & 35.42 & 43.66 & 29.00 & 47.96 & 45.78 & 29.02 & 47.30 & 42.51 \\
        SSL-ALPNet & \textbf{71.81} & 72.36 & \textbf{78.29} & 70.96 & \textbf{73.35} & 85.18 & 81.92 & \textbf{76.10} & 72.18 & 78.84 \\
        DenseMP & 68.35 & \textbf{73.55} & 76.58 & \textbf{72.88} & 72.84 & \textbf{86.28} & \textbf{82.62} & 76.04 & \textbf{74.37} & \textbf{79.83} \\
        \bottomrule
    \end{tabular}
\end{table*}

\begin{table*}[ht]
    \centering
    \caption{Comparison with sota few-shot medical image segmentation methods on setting 2.}
    \label{table:setting2}
    \begin{tabular}{l|cccc|c|cccc|c}
        \toprule
        & \multicolumn{5}{c}{Abd-CT Dice} & \multicolumn{5}{c}{Abd-MRI Dice} \\
        \cmidrule(lr){2-6} \cmidrule(lr){7-11}
        Method & RK & LK & Liver & Spleen & Mean & RK & LK & Liver & Spleen & Mean \\
        \midrule
        Vanilla PANet & 17.37 & 32.34 & 38.42 & 29.59 & 29.43 & 38.64 & 53.45 & 42.26 & 50.90 & 46.33 \\
        SE-Net & 14.34 & 32.83 & 0.27 & 0.23 & 11.91 & 61.32 & 62.11 & 27.43 & 51.80 & 50.66 \\
        SSL-ALPNet & 54.82 & 63.34 & \textbf{73.65} & 60.25 & 63.02 & 78.39 & 73.63 & \textbf{73.05} & 67.02 & 73.02 \\
        DenseMP & \textbf{64.10} & \textbf{65.95} & 73.21 & \textbf{70.30} & \textbf{68.39} & \textbf{82.78} & \textbf{79.61} & 72.71 & \textbf{72.33} & \textbf{76.86} \\
        \bottomrule
    \end{tabular}
\end{table*}

\subsection{Stage 2: Few-shot-aware Superpixel Guided Dense Pre-training}
In the first stage, the backbone network learns to generate general dense feature maps suitable for medical image segmentation through pre-training. However, the segmentation head does not benefit directly from this process. Given that our task focuses on few-shot medical image segmentation, the key aspects are "segmentation" and "few-shot". This raises the question: Can we design an unsupervised pre-training algorithm for "few-shot" as well?

To address this challenge, we need to solve two issues: 1) how to obtain the supervision signal without labels to guide the learning process? and 2) how to simulate the learning process of few-shot learning? Inspired by Ouyang et al. \cite{ouyang2020self}, we propose using superpixels as pseudo labels to tackle these problems.

Specifically, for each medical image data $I$, we first employ a traditional graph-based algorithm \cite{felzenszwalb2004efficient} to generate a superpixel map $Y$. This algorithm is fully unsupervised. In the superpixel map $Y$, there are hundreds of superpixel clusters. We randomly select one cluster as the target class and assign all other regions as background. In this way, we can obtain a pseudo-binary segmentation mask, denoted as $M_p$, for image $I$. This addresses the first issue: we can use the pseudo-binary segmentation mask as a supervision signal.

For the second issue, to simulate the few-shot learning process, we must construct an episode of data for training following the common few-shot segmentation learning scheme. To achieve this, we treat image $I$ as the support image $I_s$ and define its corresponding pseudo mask as the support mask $M_{ps}$. Then, we apply random geometric and intensity transformations, denoted as $\mathcal{T}$, to image $I$ and mask $M_p$ to obtain the query image $I_q$ and its corresponding mask $M_{pq}$:

$$
(I_q, M_{pq}) = \mathcal{T}(I, M_p)
$$

We can assume that $I_s$ and $I_q$ are medical image slices captured from different patients. This assumption is reasonable because, for the same body part, the CT or MRI scans of different patients share the same biological structure and appear similar. Therefore, the differences can be well-simulated by geometric and intensity transformations.

Through the above procedure, we obtain an episode $E = \{(I_q, M_{pq}), (I_s, M_{ps})\}$. We then apply the standard few-shot training pipeline of PA-Net \cite{wang2019panet} to train both the backbone and the segmentation head. In particular, we use the cross-entropy loss $\mathcal{L}_{CE}$ as the segmentation loss, and the prototypical alignment regularization $\mathcal{L}_{PA}$ as an auxiliary loss:

$$
\mathcal{L} = \mathcal{L}_{CE} + \lambda \mathcal{L}_{PA}
$$

Here, $\lambda$ is a trade-off parameter balancing the two loss components. We refer to this pre-training process as Few-shot-aware Superpixel Guided Dense Pre-training. After completing this stage, both the backbone network and the segmentation head can extract highly discriminative and general features, which contribute to enhancing the performance of few-shot medical image segmentation models.

In summary, the Few-shot-aware Superpixel Guided Dense Pre-training stage offers significant advantages specifically for few-shot medical image segmentation tasks by providing a highly relevant pre-trained model. Leveraging superpixels as pseudo labels greatly mitigates the need for costly and time-consuming manual annotations, which are particularly challenging to obtain in the medical domain due to the required expertise. This pre-training stage focuses on simulating the few-shot learning process, enabling the model to learn discriminative and general features more effectively. For example, when the model is applied to segment a specific organ, such as the liver, in abdominal CT scans, it is better prepared to handle variations in patient anatomies, imaging conditions, and subtle differences in liver appearance due to the initial knowledge it has gained from the pre-training stage. As a result, the pre-trained model offers improved performance and adaptability in medical image segmentation tasks like the liver segmentation example mentioned. By emphasizing the advantages in medical imaging, this stage contributes to the development of more robust and effective few-shot medical image segmentation models, ultimately benefiting real-world clinical applications.

\section{Experiments}

\begin{figure*}[htbp]
	\centering
	\includegraphics[width=0.85\textwidth]{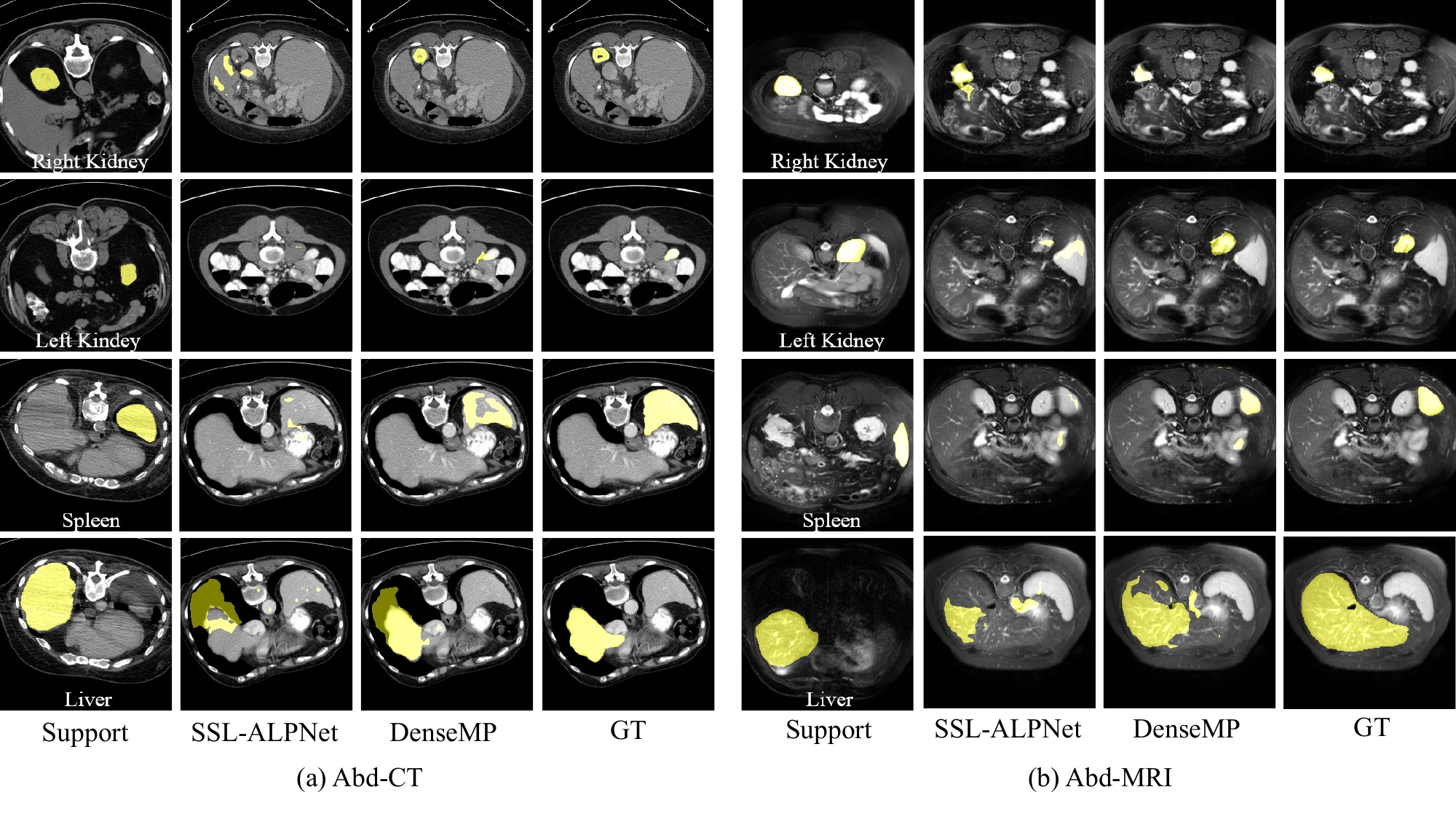}
	\caption{Qualitative comparison. (a) Visualisation results on Abd-CT dataset. (b) Visulization results on Abd-MRI dataset.}
	\label{fig:qualitative}
\end{figure*}

\subsection{Setup}

\paragraph{Datasets} In the first stage of our study, we conducted segmentation-aware dense contrastive pre-training using the ROCO dataset. Subsequently, in the second stage, we performed few-shot aware superpixel guided dense pre-training on both the abdomen CT dataset (Abd-CT) and abdomen MRI dataset (Abd-MRI). Finally, we fine-tuned the model on the Abd-CT and Abd-MRI datasets to demonstrate the final segmentation results.

The datasets employed in our study are as follows:

\begin{itemize}
    \item \textbf{ROCO} is a comprehensive, large-scale, and multi-modal medical imaging dataset \cite{pelka2018radiology}. It encompasses over 81,000 radiology images across various medical imaging modalities, including Computer Tomography, Ultrasound, X-Ray, Fluoroscopy, Positron Emission Tomography, Mammography, Magnetic Resonance Imaging, and Angiography. The diverse nature of this dataset offers a challenging environment to assess the robustness and adaptability of our pre-training model.
    
    \item \textbf{Abd-CT} is a clinically-relevant abdomen CT dataset derived from the MICCAI 2015 Multi-Atlas Abdomen Labeling challenge \cite{landman2015miccai}. It comprises 30 3D abdomen CT scans from patients presenting with a range of pathologies and intensity distribution variations between scans. This dataset poses a significant challenge due to the inherent complexity and variability in patient conditions and imaging techniques.
    
    \item \textbf{Abd-MRI} is an abdomen MRI dataset originating from the ISBI 2019 Combined Healthy Abdominal Organ Segmentation Challenge (Task 5) \cite{kavur2021chaos}. It contains 20 3D T2-SPIR MRI scans, further diversifying the imaging modalities under consideration and increasing the rigor of our study.
\end{itemize}

The data processing procedures for Abd-CT and Abd-MRI followed the same approach as described in \cite{ouyang2020self}. All images were transformed into 2D axial slices and resized to $256 \times 256$ pixels. Each 2D slice was then replicated three times in the channel dimension to fit the network. The selection of these challenging datasets, characterized by their diversity in imaging modalities and variations in patient conditions, ensures the credibility and persuasiveness of our experimental results

\paragraph{Evaluation} We adhere to the experimental settings outlined in \cite{ouyang2020self} for data pre-processing, constructing the training and testing sets, and performing 1-way 1-shot segmentation. Our study focuses on four classes of organs, which we manually divide into two groups: (\textit{spleen}, \textit{liver}) and (\textit{left kidney}, \textit{right kidney}). One group is designated for training, while the other serves as the testing set. To evaluate the robustness of our approach, we introduce two distinct experimental settings. \textbf{Setting 1} allows for the test classes to appear as background during training, whereas \textbf{Setting 2} ensures that slides containing test classes are entirely excluded from the training process, rendering test classes unseen by the network in any form. We employ standard 5-fold cross-validation to assess the performance of all methods, utilizing the Sørensen–Dice coefficient (DSC) as the evaluation metric

\paragraph{Implementation Details} In our study, we employ the PA-Net \cite{wang2019panet} as the baseline model, with ResNet101 \cite{he2016deep} serving as the backbone architecture. The input medical images have a resolution of $256 \times 256$. For the backbone network, we first pre-train it on the ImageNet \cite{deng2009imagenet} dataset for 50 epochs, followed by pre-training on the ROCO \cite{pelka2018radiology} dataset for an additional 50 epochs using segmentation-aware dense contrastive pre-training. Subsequently, we pre-train both the backbone and the segmentation head on the target datasets for 100,000 iterations. The final step involves fine-tuning the model for 5,000 iterations using both the ground truth mask and the pseudo superpixel mask. Our network implementation is conducted using the Pytorch framework.

\begin{figure*}[ht]
	\centering
	\includegraphics[width=0.8\textwidth]{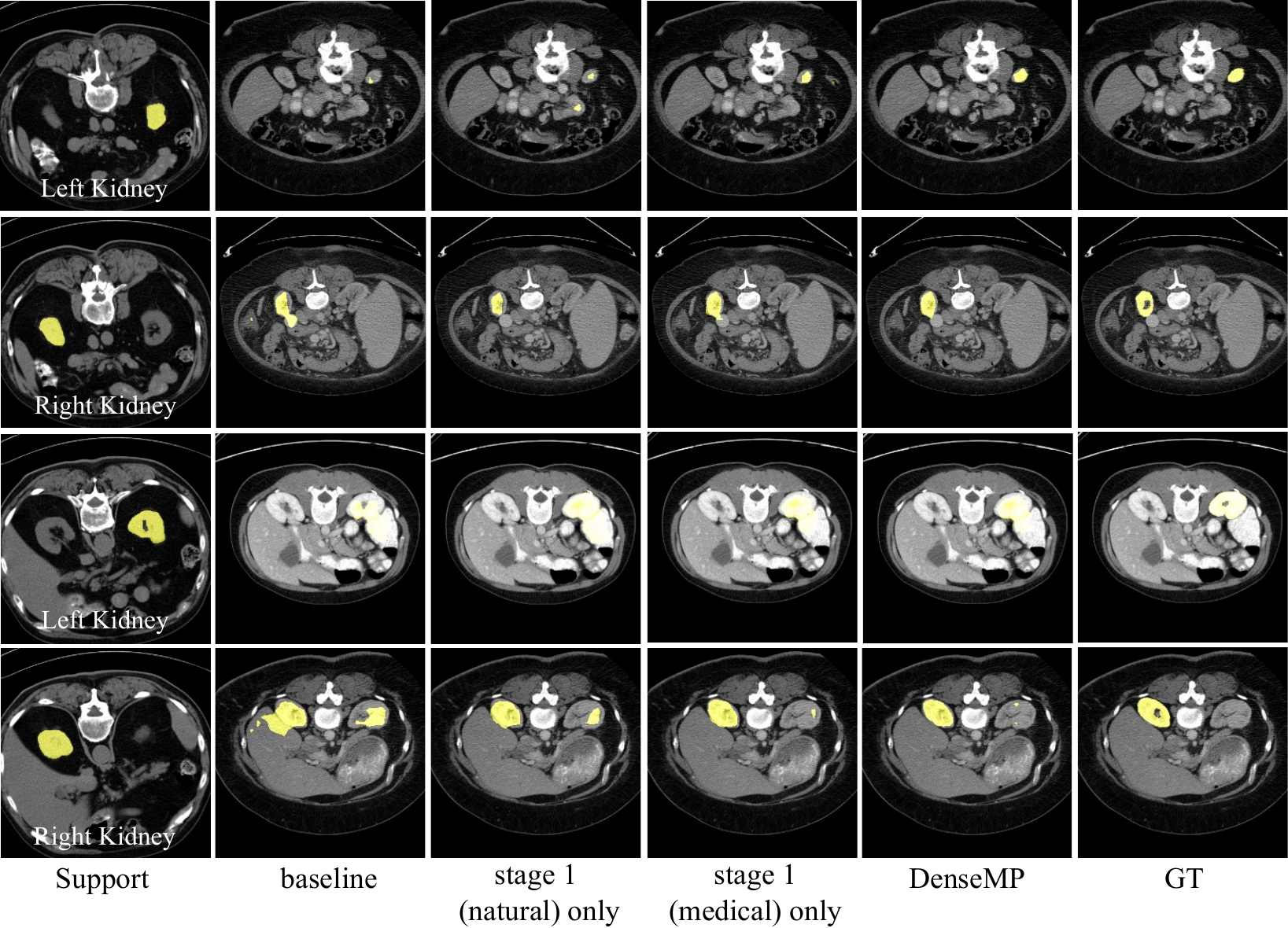}
	\caption{Visualization result of ablation study. The segmentation performance  is progressively improved with adding our proposed modules.}
	\label{fig:ablation}
\end{figure*}

\subsection{Quantitative  Results}
\noindent \textbf{Comparison with sota few-shot medical image segmentation methods:} We conduct a quantitative comparison of our proposed method with established models, including PA-Net \cite{wang2019panet}, SE-Net \cite{roy2020squeeze}, and the state-of-the-art SSL-ALPNet \cite{ouyang2020self}. The results of this comparison are presented in Tables \ref{table:setting1} and \ref{table:setting2}. In Table \ref{table:setting1}, we present the results under Setting 1. Our method, DenseMP, significantly improves the performance of the baseline model PA-Net. Specifically, for the Abd-CT dataset, the performance increases from 31.86 to 72.84, and for the Abd-MRI dataset, it improves from 38.53 to 79.83. The substantial improvement can be attributed to our two proposed pre-training stages, which aid the model in learning more general medical image-related features. As a result, DenseMP can better generalize to unseen classes during testing. Moreover, DenseMP outperforms the current state-of-the-art few-shot medical image segmentation method, SSL-ALPNet, in terms of mean Dice scores for both Abd-CT and Abd-MRI datasets, further exemplifying the superiority of our approach. Table \ref{table:setting2} displays the results under the more stringent Setting 2. Our DenseMP demonstrates a more pronounced improvement over SSL-ALPNet in comparison to the results in Table \ref{table:setting1}. For the Abd-CT dataset, DenseMP achieves a mean Dice score of 68.39, outperforming SSL-ALPNet's score of 63.02. Similarly, for the Abd-MRI dataset, DenseMP achieves a mean Dice score of 76.86, surpassing SSL-ALPNet's score of 73.02. This is due to Setting 2's condition where test classes are entirely unseen during training, as opposed to their potential presence as background in Setting 1. In such a scenario, our pre-training provides the model with more generalized cues to assist in segmenting unseen classes, preventing the model from overfitting to seen classes. In summary, the experiments conducted under Setting 2 showcase the effectiveness of our method in a more rigorous manner, highlighting the distinct advantages of our approach compared to existing methods in the field.

\noindent \textbf{Comparison with sota pretraining methods:}  To further demonstrate the superiority of our customized "pre-training" approach for few-shot medical image segmentation, we present a comparison between our method and three state-of-the-art pre-training methods: Swin-SimMIM \cite{xie2022simmim}, ViT-MAE \cite{he2022masked}, and Res50-SimCLR \cite{chen2020simple} in Table \ref{table:pretraining_setting1} and Table \ref{table:pretraining_setting2}. These methods have been proven to exhibit exceptional performance in natural image segmentation tasks.

\noindent \textbf{Comparison with sota pretraining methods:}


\begin{table*}[ht]
    \centering
    \caption{Comparison with SOTA pertaining methods on setting 1.}
   \label{table:pretraining_setting1}
   \scalebox{0.65}{
        \begin{tabular}{l|cccc|c|cccc|c}
            \toprule
            & \multicolumn{5}{c}{Abd-CT Dice} & \multicolumn{5}{c}{Abd-MRI Dice} \\
            \cmidrule(lr){2-6} \cmidrule(lr){7-11}
            Method & RK & LK & Liver & Spleen & Mean  & RK & LK & Liver & Spleen & Mean \\
            \midrule
            Swin-SimMIM & 34.81 $\pm$ 14.53 & 30.52 $\pm$ 7.00 & 68.50 $\pm$ 2.00 & 51.05 $\pm$ 6.40 & 46.22 $\pm$ 7.48 & 51.59 $\pm$ 5.93 & 50.50 $\pm$ 9.37 & 68.30 $\pm$ 4.50 & 55.91 $\pm$ 8.01 & 56.57 $\pm$ 6.95 \\
            ViT-MAE & 36.94 $\pm$ 12.67 & 33.71 $\pm$ 5.27 & 76.26 $\pm$ 5.00 & 48.58 $\pm$ 6.84 & 48.87 $\pm$ 7.44 & 56.25 $\pm$ 12.29 & 50.86 $\pm$ 11.08 & 74.73 $\pm$ 3.26 & 60.76 $\pm$ 10.43 & 60.65 $\pm$ 9.27 \\
            Res50-SimCLR & 22.42 $\pm$ 10.82 & 17.59 $\pm$ 8.02 & 55.51 $\pm$ 6.95 & 21.18 $\pm$ 8.53 & 29.18 $\pm$ 8.58 & 32.89 $\pm$ 6.12 & 27.67 $\pm$ 6.67 & 67.85 $\pm$ 1.98 & 43.17 $\pm$ 11.04 & 42.9 $\pm$ 6.45 \\
            DenseMP & \textbf{68.35 $\pm$ 10.49} & \textbf{73.56 $\pm$ 9.23} & \textbf{76.58 $\pm$ 5.15} & \textbf{72.88 $\pm$ 4.60} & \textbf{72.84 $\pm$ 7.39} & \textbf{86.28 $\pm$ 2.77} & \textbf{82.62 $\pm$ 4.43} & \textbf{76.04 $\pm$ 3.29} & \textbf{74.37 $\pm$ 8.38} & \textbf{79.83 $\pm$ 4.72} \\
            \bottomrule
        \end{tabular}
    }
\end{table*}

\begin{table*}[ht]
    \centering
       \caption{Comparison with SOTA pertaining methods on setting 2.}
    \label{table:pretraining_setting2}
    \scalebox{0.65}{
        \begin{tabular}{l|cccc|c|cccc|c}
            \toprule
            & \multicolumn{5}{c}{Abd-CT Dice} & \multicolumn{5}{c}{Abd-MRI Dice} \\
            \cmidrule(lr){2-6} \cmidrule(lr){7-11}
            Method & RK & LK & Liver & Spleen & Mean & RK & LK & Liver & Spleen & Mean \\
            \midrule
            Swin-SimMIM & 32.08 $\pm$ 16.04 & 24.89 $\pm$ 6.49 & 64.84 $\pm$ 1.87 & 45.49 $\pm$ 7.65 & 41.83 $\pm$ 8.01  & 51.25 $\pm$ 4.54 & 51.06 $\pm$ 6.14 & 67.5 $\pm$ 3.86 & 52.73 $\pm$ 9.86 & 55.64 $\pm$ 6.10 \\
            ViT-MAE & 44.85 $\pm$ 8.12  & \textbf{75.64 $\pm$ 5.40} & 70.24 $\pm$ 6.58 & 40.85 $\pm$ 7.55 & 57.90 $\pm$ 6.91 & 54.53 $\pm$ 15.13 & 49.33 $\pm$ 16.02 & \textbf{74.32 $\pm$ 3.57} & 57.57 $\pm$ 9.02 & 58.94 $\pm$ 10.93 \\
            Res50-SimCLR & 19.63 $\pm$ 10.95 & 14.49 $\pm$ 5.19 & 54.19 $\pm$ 4.42 & 20.96 $\pm$ 7.43 & 27.32 $\pm$ 7.00 & 28.93 $\pm$ 21.02 & 21.02 $\pm$ 5.08 & 62.58 $\pm$ 3.21 & 39.1 $\pm$ 13.19 & 37.91 $\pm$ 8.05 \\
            DenseMP & \textbf{64.10 $\pm$ 6.87} & 65.95 $\pm$ 5.82 & \textbf{73.21 $\pm$ 3.81} & \textbf{70.30 $\pm$ 8.35} & \textbf{68.39 $\pm$ 6.21} & \textbf{82.78 $\pm$ 2.50} & \textbf{79.61 $\pm$ 4.29} & 72.71 $\pm$ 2.99 & \textbf{72.33 $\pm$ 8.68} & \textbf{76.86 $\pm$ 4.62} \\
            \bottomrule
        \end{tabular}
    }
\end{table*}

\subsection{Qualitative  Results}

In Fig. \ref{fig:qualitative}, we present visualizations of the segmentation results obtained using our DenseMP approach in comparison to those produced by SSL-ALPNet. A close examination of these visualizations reveals several key advantages of our method over the competing approach, which can be primarily attributed to two critical components of our method: (1) segmentation-aware dense contrastive pre-training, and (2) few-shot-aware superpixel guided dense pre-training. Firstly, as evident from the first row, DenseMP is capable of accurately identifying and localizing the target organ, whereas SSL-ALPNet fails to do so and instead segments several unrelated clusters. This superior precision in detecting regions of interest can be attributed to the segmentation-aware dense contrastive pre-training, which encourages the model to learn discriminative features by comparing dense pixel-level similarities across images. Secondly, as shown in the second and fourth rows, DenseMP effectively segments a more complete representation of the target organ, whereas SSL-ALPNet is only able to identify a small portion of it. This enhanced capability in capturing the entire organ structure can be ascribed to the few-shot-aware superpixel guided dense pre-training, which leverages superpixel information to guide the learning process, thus enabling the model to generate more accurate and coherent segmentation results. In the third row, DenseMP successfully locates and segments the target organ, while SSL-ALPNet appears to be focused on an incorrect organ. This further emphasizes the robustness of our method in correctly distinguishing between different organs. The combination of segmentation-aware dense contrastive pre-training and few-shot-aware superpixel guided dense pre-training allows our model to effectively differentiate between various organ structures, leading to improved segmentation performance. In summary, our proposed DenseMP approach significantly outperforms SSL-ALPNet in terms of few-shot segmentation accuracy, which can be attributed to the effectiveness of our unsupervised pre-training pipeline, consisting of the segmentation-aware dense contrastive pre-training and the few-shot-aware superpixel guided dense pre-training. 

\begin{figure}[t]
	\centering
	\includegraphics[width=0.9\linewidth]{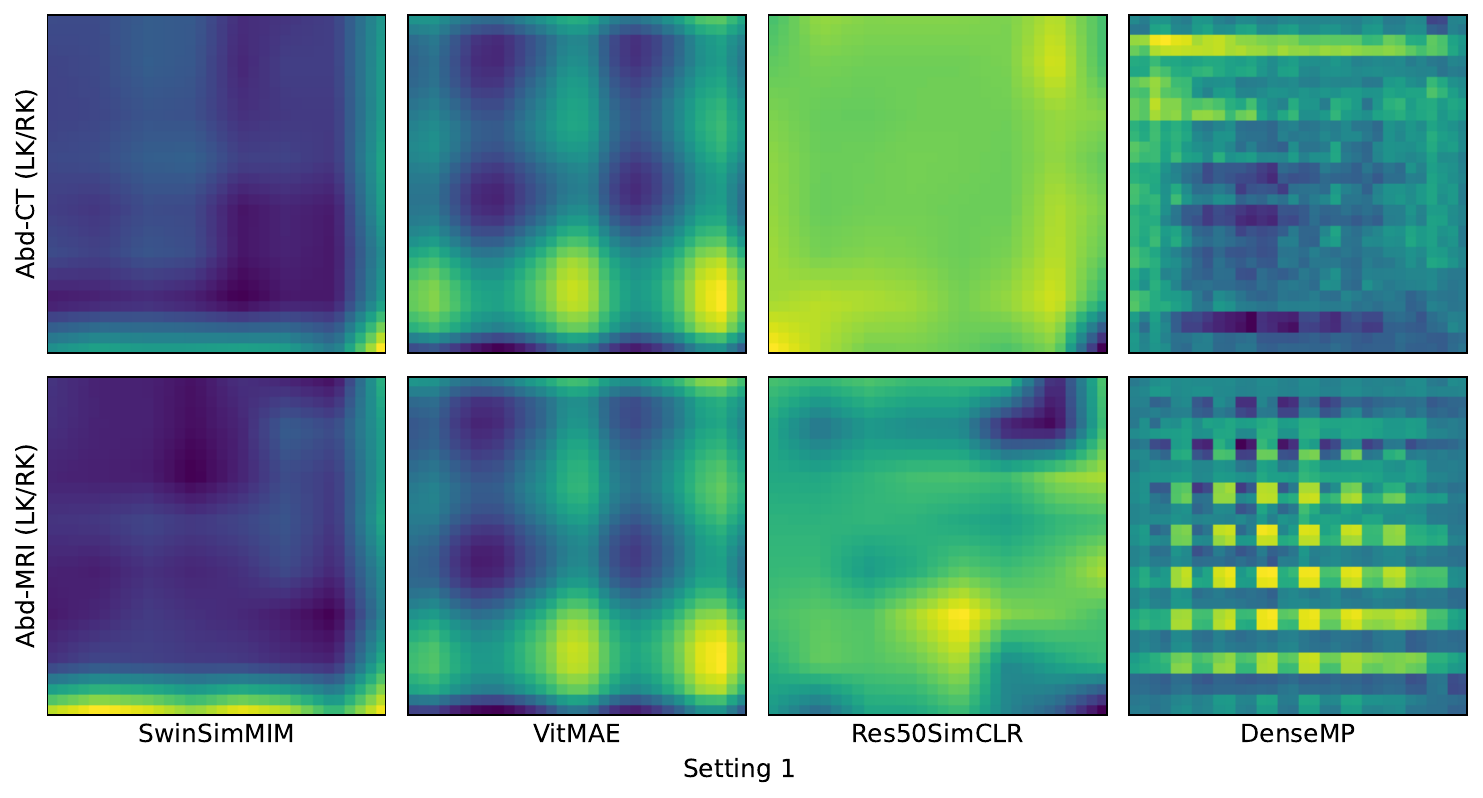}
	\caption{Visualization of learned features on Setting 1}
	\label{fig:feature_setting1}
\end{figure}

\begin{figure}[t]
	\centering
	\includegraphics[width=0.9\linewidth]{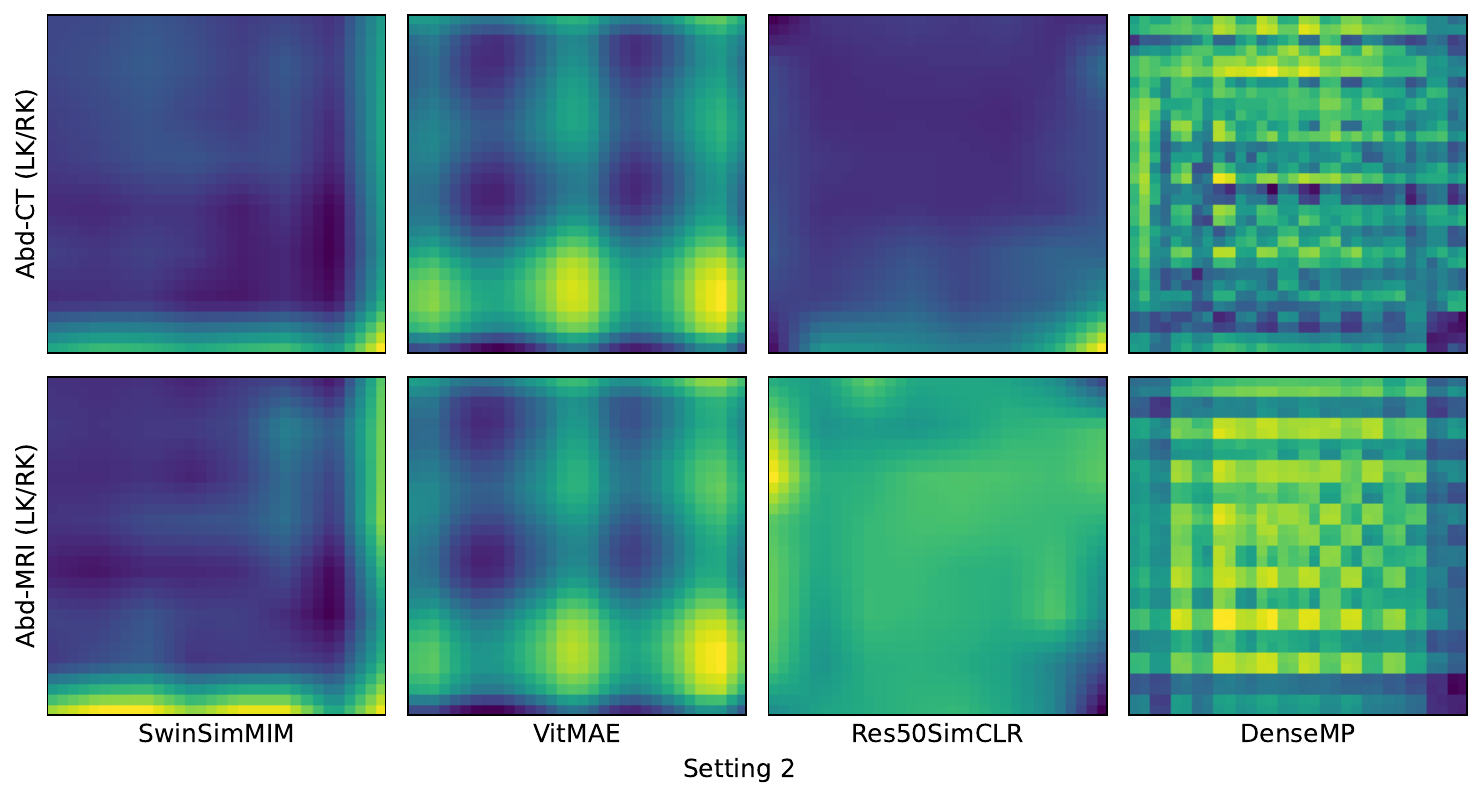}
	\caption{Visualization of learned features on Setting 2}
	\label{fig:feature_setting2}
\end{figure}

Furthermore, as previously mentioned, our method outperforms other pre-training approaches, raising the question of what features our method has learned to achieve this superior performance. In Fig. \ref{fig:feature_setting1} and Fig. \ref{fig:feature_setting2}, we visualize the learned features of DenseMP, Swin-SimMIM, ViT-MAE, and Res50-SimCLR. From these two figures, it can be observed that the features learned by Res50-SimCLR exhibit uniform activation across all pixel points, which is due to the global nature of its pre-training. In contrast, both Swin-SimMIM and ViT-MAE demonstrate some degree of dense activation, with different pixel points exhibiting varying activation levels and a certain degree of clustering characteristics. DenseMP, when compared to the other methods, displays densely packed, small block-shaped activations in its feature maps. This suggests that the features are indeed densely localized responses, making them well-suited for few-shot segmentation tasks. By employing a dense contrastive pre-training strategy, DenseMP is able to capture rich, localized information from input images. This dense feature representation allows the model to better differentiate between various fine-grained structures and textures, which is critical for few-shot segmentation tasks where limited labeled data is available. Consequently, this dense and localized feature representation contributes to the superior performance of DenseMP in few-shot medical image segmentation.

\subsection{Ablation Study}
\begin{table}[htbp]
	\centering
	\begin{tabular}{p{4.0cm}|p{0.8cm}<{\centering}|p{0.8cm}<{\centering}|p{0.8cm}<{\centering}}
		\hline
		\multirow{2}{*}{Method} & \multicolumn{3}{c}{Abd-CT Dice} \\ \cline{2-4}
		& RK & LK & Mean \\ \hline
		baseline & 40.34 & 47.13 & 43.73 \\ \hline
        stage 1 only (natural) & 48.52 & 49.08 & 48.80 \\ \hline
        stage 1 only (medical) & 51.79 & 54.53 & 53.16 \\ \hline
        stage 1 + stage 2 (DenseMP) & \textbf{56.91} & \textbf{57.48} & \textbf{57.20} \\ \hline
	\end{tabular}
	\caption{Ablation experiments results using \textbf{setting 2} on \textbf{Abd-CT} Fold 1 for kidneys. \textbf{RK} means right kidney. \textbf{LK} means left kidney.  Best results are in \textbf{bold}. } 
	\label{table:ablation}
\end{table}

In this section, we investigate the impact of our proposed dual pre-training modules through a series of ablation studies, conducted on the Abd-CT dataset. We specifically choose to adopt the experimental Setting 2, as it ensures that slides containing test classes are entirely excluded from the training process, rendering test classes unseen by the network in any form. This setting not only presents a challenging scenario, but also reflects a more realistic and clinically relevant situation for practical applications.

\noindent \textbf{Effect of Segmentation-Aware Dense Contrastive Pre-Training Stage.} First, we assess the effectiveness of the segmentation-aware dense contrastive pre-training module, focusing on its ability to enhance the backbone's capability to learn more general features. To ensure efficiency, our experiments are conducted on Abd-CT Fold 1, with the results presented in Table \ref{table:ablation}. In the row "Stage 1 (natural)", we employ the segmentation-aware dense contrastive pre-training module to pre-train our network, utilizing only the ImageNet dataset—a large-scale natural image dataset. The results indicate a substantial improvement in the performance of our few-shot segmentation model, particularly for the right kidney. Subsequently, we incorporate the ROCO medical image dataset for continued pre-training using Stage 1. As the third row demonstrates, few-shot segmentation performance for both the right and left kidneys is significantly enhanced compared to pre-training the backbone solely on natural images. This improvement can be attributed to the unique and general characteristics of medical images, which are absent in natural images, being learned from the ROCO dataset.

\noindent  \textbf{Effect of few-shot-aware superpixel guided dense pre-training stage.}  As shown in Table \ref{table:ablation}, comparing the performance of the model with Stage 1 only (medical) pre-training (row 4) and the model with both Stage 1 and Stage 2 (DenseMP) pre-training strategies (row 6), we observe a significant improvement in the Dice scores for the right kidney (RK), left kidney (LK), and mean. The scores increase from 51.79, 54.53, and 53.16, respectively, to 56.91, 57.48, and 57.20, respectively. This supports our hypothesis that incorporating the few-shot-aware superpixel guided dense pre-training stage (Stage 2) is beneficial. The underlying reason for this enhancement is the simulation of the few-shot segmentation process using superpixels in Stage 2. This approach enables the model to learn an effective initialization for few-shot learning, as it mimics the natural scarcity of labeled data in a real-world few-shot scenario. By tailoring Stage 2 to few-shot learning, we can effectively capture the unique challenges and characteristics of few-shot segmentation tasks in the medical domain. Moreover, the combination of Stage 1 and Stage 2 pre-training strategies exhibits a synergistic effect, leading to a model that is better equipped to handle few-shot medical image segmentation tasks. This finding highlights the importance of designing pre-training strategies specifically tailored for few-shot learning in the context of medical image segmentation.

\noindent  \textbf{Effect of dense information in segmentation-aware dense contrastive pre-training stage.} To evaluate the effectiveness of dense information, we conducted an ablation study with the same setup but substituting the dense pre-training method with SimCLR, a global contrastive learning method. In Table \ref{table:ablation2}, we present the results of this comparison to investigate the impact of dense pre-training versus global contrastive learning for few-shot medical image segmentation. The results in Table \ref{table:ablation2} demonstrate that although the global contrastive learning pre-training method (SimCLR) improves the performance of the baseline model SSL-ALPNet, the improvement is less significant compared to the dense pre-training method (DenseMP). Specifically, the Dice scores for the right kidney (RK), left kidney (LK), and mean increase from 54.82, 63.34, and 59.08 for SSL-ALPNet to 61.73, 63.43, and 62.58 for SimCLR. However, the DenseMP (Ours) method further improves the performance, achieving the best Dice scores of 64.10, 65.95, and 65.03, respectively.
These findings verify that dense information is more essential for few-shot medical image segmentation than global information. The superior performance of the DenseMP method highlights the importance of leveraging dense information during the pre-training stage, particularly for tasks related to few-shot medical image segmentation.

In Figure \ref{fig:ablation}, we provide a visual representation of the segmentation results to further illustrate the impact of the dense information in the segmentation-aware dense contrastive pre-training stage (Stage 1) and the few-shot-aware superpixel guided dense pre-training stage (Stage 2) on few-shot medical image segmentation performance. As observed in Figure \ref{fig:ablation}, the segmentation performance is progressively improved by incorporating our proposed modules.

\begin{table}[htbp]
	\centering
	\begin{tabular}{p{3.7cm}|p{0.8cm}<{\centering}|p{0.8cm}<{\centering}|p{0.8cm}<{\centering}}
		\hline
		\multirow{2}{*}{Method} & \multicolumn{3}{c}{Abd-CT Dice} \\ \cline{2-4}
		& RK & LK & Mean \\ \hline
		SSL-ALPNet & 54.82 & 63.34 & 59.08 \\ \hline
        DenseMP (SimCLR) & 61.73 & 63.43 & 62.58 \\ \hline
        DenseMP (Ours) & \textbf{64.10} & \textbf{65.95} & \textbf{65.03} \\ \hline
	\end{tabular}
	\caption{Ablation experiments results using \textbf{setting 2} on \textbf{Abd-CT} for kidneys. \textbf{RK} means right kidney. \textbf{LK} means left kidney.  Best results are in \textbf{bold}.} 
	\label{table:ablation2}
\end{table}

\begin{table}[htbp]
	\centering
	\begin{tabular}{p{3.7cm}|p{0.8cm}<{\centering}|p{0.8cm}<{\centering}|p{0.8cm}<{\centering}}
		\hline
		\multirow{2}{*}{Min size(px)} & \multicolumn{3}{c}{Abd-CT Dice} \\ \cline{2-4}
		& RK & LK & Mean \\ \hline
		Avg. Size in 2D (px) & 798 & 799 &  \\ \hline
		100 & 56.45 & 55.27 & 53.97 \\ \hline
        400 & \textbf{64.10} &\textbf{65.95}& \textbf{65.03} \\ \hline
        1600 & 53.12 & 51.26 & 52.19 \\ \hline
	\end{tabular}
	\caption{Ablation experiments results using \textbf{setting 2} on \textbf{Abd-CT} for kidneys. \textbf{RK} means right kidney. \textbf{LK} means left kidney.  Best results are in \textbf{bold}.} 
	\label{table:ablation3}
\end{table}

\noindent  \textbf{Effect of minimum pseudolabel sizes in few-shot-aware superpixel guided dense pre-training stage.}To investigate the influence of varying minimum pseudolabel sizes generated during the data processing procedure of pre-training Stage 2, we conducted a series of experiments. The objective of this analysis is to prevent the model's training from being misled by excessively small pseudolabels and to identify the optimal number of superpixels for this stage. Table \ref{table:ablation3} presents the results of these experiments, revealing that the ideal number of superpixels should be neither too large nor too small. Based on our findings, setting the minimum pseudolabel size to 400 yields the best performance in terms of few-shot medical image segmentation. The optimal choice of 400 can be attributed to several factors, including the balance between granularity and noise, spatial coherence, computational efficiency, and robustness to class imbalance. This value ensures a fine-grained representation of structures without introducing noise, maintains spatial coherence for capturing local contextual information, provides a reasonable trade-off between computational demands and segmentation performance, and mitigates the effects of class imbalance by promoting diverse learning from various structures and regions.

\section{Conclusion}
In this study, we introduce DenseMP, a novel two-stage approach for few-shot medical image segmentation. DenseMP combines segmentation-aware dense contrastive pre-training and few-shot-aware superpixel guided dense pre-training, facilitating unsupervised learning of valuable features. Our work features extensive comparative experiments and ablation studies, which collectively validate the effectiveness and rationality of each component in our proposed method.

One limitation of our current approach is its reliance on pre-defined superpixels, which might not always optimally represent the underlying structures in medical images. In future work, we plan to explore adaptive superpixel generation techniques that better capture the intricate details of various anatomical structures. Additionally, we aim to investigate the potential of incorporating multi-modal medical imaging data to further enhance the performance of DenseMP.

\bibliography{aaai24}

\end{document}